\documentclass[11pt,a4paper]{article}
\usepackage[hyperref]{emnlp2020}
\usepackage{times}
\usepackage{latexsym}

\usepackage{wasysym}

\usepackage{microtype}
\usepackage{graphicx}
\usepackage{multirow}
\usepackage{makecell}

\aclfinalcopy 

\title{InfoMiner at WNUT-2020 Task 2: Transformer-based Covid-19 Informative Tweet Extraction}

\author{Hansi Hettiarachchi$^\heartsuit$, Tharindu Ranasinghe$^\ddagger$ \\

  $^\heartsuit$School of Computing and Digital Technology, Birmingham City University, UK \\
    
  $^\ddagger$Research Group in Computational Linguistics, University of Wolverhampton, UK \\

  {\tt hansi.hettiarachchi@mail.bcu.ac.uk } \\
  {\tt tharindu.ranasinghe@wlv.ac.uk }} 

\date{}

\begin{document}
\maketitle
\begin{abstract}
Identifying informative tweets is an important step when building information extraction systems based on social media. WNUT-2020 Task 2 was organised to recognise informative tweets from noise tweets. In this paper, we present our approach to tackle the task objective using transformers. Overall, our approach achieves 10$^{th}$ place in the final rankings scoring 0.9004 F1 score for the test set.
\end{abstract}

\section{Introduction} \label{sec:introduction}
By 31st August 2020, coronavirus COVID-19 is affecting 213 countries around the world infecting more than 25 million people and killing more than 
800,000. Recently, much attention has been given to build monitoring systems to track the outbreaks of the virus. However, due to the fact that most of the official news sources update the outbreak information only once or twice a day, these monitoring tools have begun to use social media as the medium to get information. 

 There is a massive amount of data on social networks, e.g. about 4 millions of COVID-19 English tweets daily on the Twitter platform. However, majority of these tweets are uninformative. Thus it is important to be able to select the informative ones for downstream applications. Since the manual approaches to identify the informative tweets require significant human efforts, an automated technique to identify the informative tweets will be invaluable to the community. 
 
 The objective of this shared task is to automatically identify whether a COVID-19 English tweet is informative or not. Such informative Tweets provide information about recovered, suspected, confirmed and death cases as well as location or travel history of the cases. The participants of the shared task were required to provide predictions for the test set provided by the organisers whether a tweet is informative or not. Our team used recently released transformers to tackle the problem. Despite achieving 10$^{th}$ place out of 55 participants and getting high evaluation score, our approach is simple and efficient. In this paper we mainly present our approach that we used in this task. We also provide important resources to the community: the code, and the trained classification models will be freely available to everyone interested in working on identifying informative tweets using the same methodology \footnote{The GitHub repository is publicly available on \url{https://github.com/hhansi/informative-tweet-identification}}.

\section{Related Work} \label{sec:related-work}
In the last few years, there have been several studies published on the application of computational
methods in order to identify informative contents from tweets. Most of the earlier methods were based on traditional machine learning models like logistic regression and support vector machines with heavy feature engineering. \citet{10.1145/1963405.1963500} investigate tweet newsworthiness classification using features representing the message, user, topic
and the propagation of messages. Others use features based on social influence, information propagation, syntactic and combinations of local linguistic features as well as user history and user opinion to select informative tweets \cite{6113128,10.1145/2009916.2009954,ICWSM136057}. Due to the fact that training set preparation is difficult when it comes informative tweet identification, several studies suggested unsupervised methods. \citet{10.1145/1653771.1653781} built a news processing system, called \textit{TwitterStand} using an unsupervised approach to classify tweets collected from pre-determined users who frequently post news about events. Even though these traditional approaches have provided good results, they are no longer the state of the art.

Considering the recent research, there was a tendency to use deep learning-based methods to identify informative tweets since they performed better than traditional machine learning-based methods. To mention few, \citet{alrashdi2019deep} suggested an approach based on Bidirectional Long Short-Term Memory (Bi-LSTM) models trained using word embeddings. Another research proposed a deep multi-modal neural network based on images and text in tweets to recognise informative tweets \cite{Kumar2020}. Among the different neural network models available, transformer models received a huge success in the area of natural language processing (NLP) recently. Since the release of BERT \cite{devlin-etal-2019-bert}, transformer models gained a wide attention of the community and they were successfully applied for wide range of tasks including tweet classification tasks such as offensive tweet identification \cite{ranasinghe2019brums} and topic identification \cite{yuksel2019turkish}. But we could not find any previous work on transformers for informative tweet classification. Hence, we decided to use transformer for our approach and this study will be important to the community.  

\section{Task Description and Data Set} \label{sec:task-desscription}

WNUT-2020 Task 2: Identification of informative COVID-19 English Tweets \cite{covid19tweet} is to develop a system which can automatically categorise the tweets related to coronavirus as informative or not. A data set of 10K tweets which are labelled as \textit{informative} and \textit{uninformative} is released to conduct this task. The class distributions of the data set splits are mentioned in Table \ref{table:data-set}. 

\begin{table}[h]
\centering
\begin{tabular}{l|l|l}
\hline
\textbf{Data set} & \textbf{Informative} & \textbf{Uninformative}\\
\hline
Training & 3303 & 3697 \\
Validation & 472 & 528 \\
Test & 944 & 1056 \\
\hline
\end{tabular}
\caption{\label{table:data-set}
Class distribution of data set splits
}
\end{table}

\section{Methodology}
The motivation behind our methodology is the recent success that the transformers had in wide range of NLP tasks like language generation  \cite{devlin-etal-2019-bert}, sequence classification \cite{ranasinghe-etal-2020-brums,ranasinghe2019brums,ranasinghe-etal-2020-multilingual}, word similarity \cite{hettiarachchi-etal-2020-brums}, named entity recognition \cite{10.1145/3394486.3403149} and question and answering \cite{yang-etal-2019-end}. The main idea of the methodology is that we train a classification model with several transformer models in-order to identify informative tweets.

\subsection{Transformers for Text Classification}
\label{subsec:classification}
Predicting whether a certain tweet is informative or not can be considered as a sequence classification task. Since the transformer architectures have shown promising results in sequence classification tasks \cite{ranasinghe-etal-2020-brums,ranasinghe2019brums,ranasinghe-etal-2020-multilingual}, the basis for our methodology was transformers. Transformer architectures have been trained on general tasks like language modelling and then can be fine-tuned for classification tasks. \cite{10.1007/978-3-030-32381-3_16}

\begin{figure}[ht]
\centering
\includegraphics[scale=0.41]{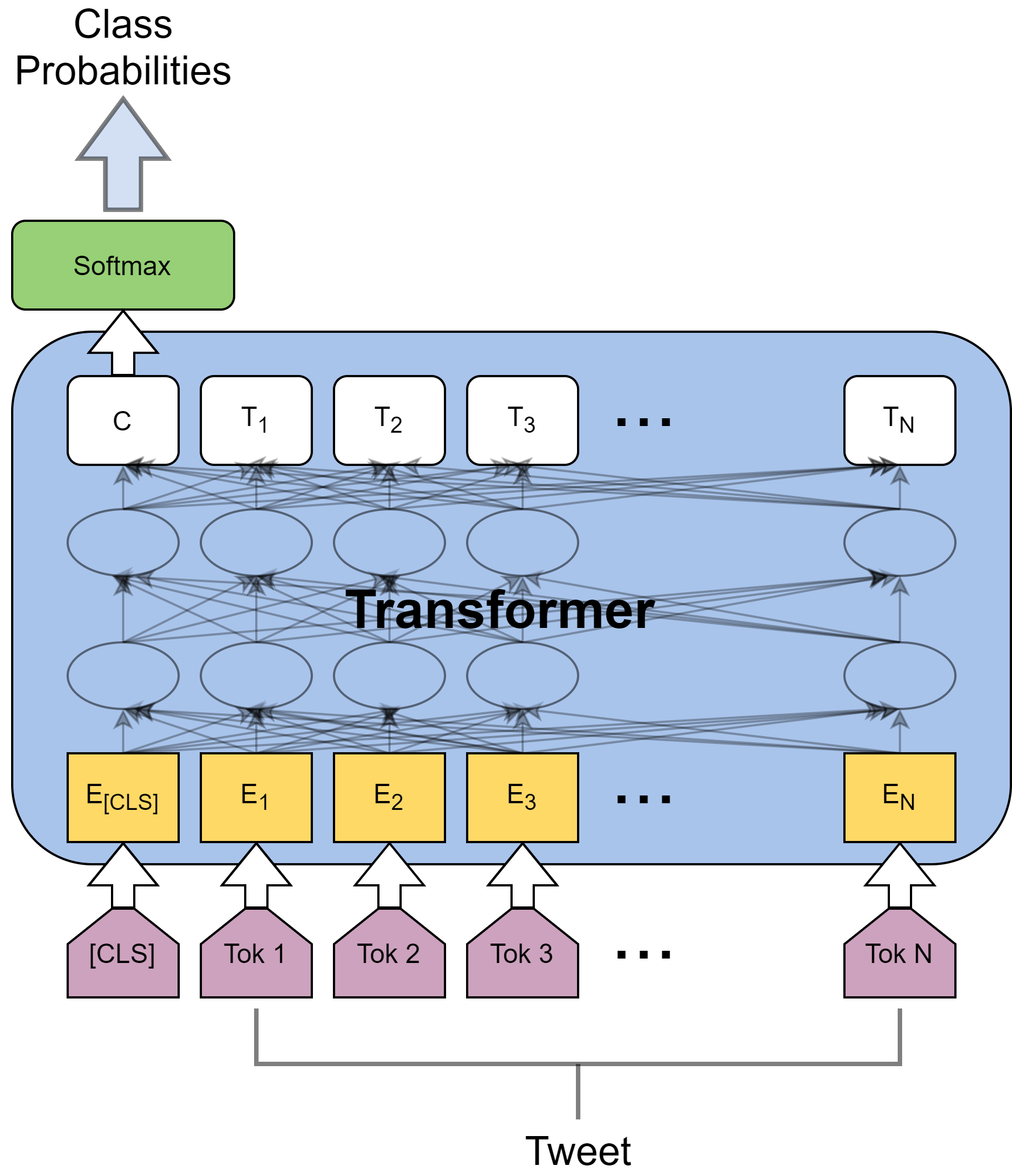}
\caption{Text Classification Architecture}
\label{fig:architecture}
\end{figure}

Transformer models take an input of a sequence and outputs the representations of the sequence. There can be one or two segments in a sequence which are separated by a special token [SEP]. In this approach we considered a tweet as a sequence and no [SEP] token is used. Another special token [CLS] is used as the first token of the sequence which contains a special classification embedding. For text classification tasks, transformer models take the final hidden state $\textbf{h}$ of the [CLS] token as the representation of the whole sequence \cite{10.1007/978-3-030-32381-3_16}. A simple softmax classifier is added to the top of the transformer model to predict the probability of a class $c$ as shown in Equation \ref{equ:softmax} where $W$ is the task-specific parameter matrix. The architecture of transformer-based sequence classifier is shown in Figure \ref{fig:architecture}.

\begin{equation}
\label{equ:softmax}
p(c|\textbf{h}) = softmax(W\textbf{h}) 
\end{equation}

\subsection{Transformers}
\label{subsec:transformers}
We used several pre-trained transformer models in this task. These models were used mainly considering the popularity of them (e.g.\ BERT \cite{devlin-etal-2019-bert}, XLNet \cite{yang2019xlnet}, RoBERTa \cite{liu2019roberta}, ELECTRA \cite{clark2020electra}, ALBERT \cite{Lan2020ALBERT}) and relatedness to the task (e.g.\ COVID-Twitter-BERT (CT-BERT) \cite{muller2020covid} and BERTweet \cite{BERTweet}).

BERT \cite{devlin-etal-2019-bert} was the first transformer model that gained a wide attention of the NLP community. It proposes a masked language modelling (MLM) objective, where some of the tokens of a input sequence are randomly masked, and the objective is to predict these masked positions taking the corrupted sequence as input. As we explained before BERT uses special tokens to obtain a single contiguous sequence for each input sequence. Specifically, the first token is always a special classification token [CLS] which is used for sentence-level tasks. 

RoBERTa \cite{liu2019roberta}, ELECTRA \cite{clark2020electra} and ALBERT \cite{Lan2020ALBERT} can all be considered as variants of BERT. They make a few changes to the BERT model and achieves
substantial improvements in some NLP tasks \cite{liu2019roberta,clark2020electra,Lan2020ALBERT}. XLNet on the other hand takes a different approach to BERT \cite{yang2019xlnet}. XLNet proposes a new auto-regressive method based on permutation language modelling (PLM) \cite{10.5555/2946645.3053487} without introducing any new symbols such as [MASK] in BERT. Also there are significant changes in the XLNet architecture like adopting two-stream self-attention and Transformer-XL \cite{dai-etal-2019-transformer}. Due to this XLNet outperforms BERT in multiple NLP downstream tasks \cite{yang2019xlnet}.

We also used two transformer models based on Twitter; CT-BERT and BERTweet. The CT-BERT model is based on the BERT-LARGE model and trained on a corpus of 160M tweets about the coronavirus \cite{muller2020covid} while the BERTweet model is based on BERT-BASE model and trained on general tweets \cite{BERTweet}.

\begin{table*}[!htbp]
\centering
\begin{tabular}{l|ccc|ccc}
\hline
\textbf{Strategy} & \multicolumn{3}{c|}{Single-model} & \multicolumn{3}{c}{MSE (N=3)} \\
\hline
\multicolumn{1}{l|}{\textbf{Model}} & \textbf{P} & \textbf{R} & \textbf{F1} & \textbf{P} & \textbf{R} & \textbf{F1}\\
\hline
\textit{bert-large-cased} & 0.9031 & 0.8686 & 0.8855 & 0.8884 & 0.8941 & 0.8912 \\
\textit{roberta-large} & 0.9056 & 0.8941 & 0.8998 & 0.8926 & 0.9153 & 0.9038 \\
\textit{albert-xxlarge-v1} & 0.9009 & 0.8856 & 0.8932 & 0.9032 & 0.8898 & 0.8965 \\
\textit{xlnet-large-cased} & 0.8778 & 0.9280 & 0.9022 & 0.8743 & 0.9280 & 0.9003 \\
\textit{electra-large-generator} & 0.8297 & 0.8771 & 0.8527 & 0.8901 & 0.8750 & 0.8825 \\
\textit{bertweet-base} & 0.8710 & 0.8968 & 0.8753 & 0.8741 & 0.8998 &  0.8780 \\
\textit{covid-twitter-bert} & 0.8984 & 0.9364 & 0.9170 & 0.9002 & 0.9364 & 0.9180 \\
\hline
\end{tabular}
\caption{\label{table:transformer-models}
Results of different transformer models (All these experiments are executed for 3 learning epochs with $1e^{-5}$ learning rate.)
}
\end{table*}

\subsection{Data Preprocessing}
\label{subsec:data_preprocessing}
Few general data preprocessing techniques were employed with InfoMiner to preserve the universality of this method. More specifically, used techniques can be listed as removing or filling usernames and URLs, and converting emojis to text. Further, for uncased pretrained models (e.g. \textit{albert-xxlarge-v1}), all tokens were converted to lower case. 

In WNUT-2020 Task 2 data set, mention of a user is represented by \textit{@USER} and a URL is represented by \textit{HTTPURL}. For all the models except CT-BERT and BERTweet, we removed those mentions. The main reason behind this step is to remove noisy text from data. CT-BERT and BERTweet models are trained on tweet corpora and usernames and URLs are introduced to the models using special fillers. CT-BERT model knows a username as \textit{twitteruser} and URL as \textit{twitterurl}. Likewise, BERTweet model used the filler \textit{@USER} for usernames and \textit{HTTPURL} for URLs. Therefore, for these two models we used the corresponding fillers to replace usernames and URLs in the data set. 

Emojis are found to play a key role in expressing emotions in the context of social media \cite{hettiarachchi2019emoji}. But, we cannot assure the existence of embeddings for emojis in pretrained models. Therefore as another essential preprocessing step, we converted emojis to text. For this conversion we used the Python libraries \textit{demoji}\footnote{demoji repository - \url{https://github.com/bsolomon1124/demojis}} and \textit{emoji}\footnote{emoji repository - \url{https://github.com/carpedm20/emoji}}. \textit{demoji} returns a normal descriptive text and \textit{emoji} returns a specifically formatted text. For an example, the conversion of \smiley \space is `slightly smiling face' using \textit{demoji} and `:slightly\textunderscore smiling\textunderscore face:' using \textit{emoji}. For all the models except CT-BERT and BERTweet, we used \textit{demoji} supported conversion. For CT-BERT and BERTweet \textit{emoji} supported conversion is used, because these models are trained on correspondingly converted Tweets. 

\begin{table*}[!htbp]
\centering
\begin{tabular}{l|l|ccc|ccc|ccc}
\hline
\multicolumn{2}{l|}{\textbf{Learning R.}}  & \multicolumn{3}{c|}{$1e^{-5}$} & \multicolumn{3}{c|}{$1e^{-6}$} & \multicolumn{3}{c}{$2e^{-5}$} \\
\hline
\textbf{S. 1} & \textbf{S. 2} & \textbf{P} & \textbf{R} & \textbf{F1} & \textbf{P} & \textbf{R} & \textbf{F1} & \textbf{P} & \textbf{R} & \textbf{F1} \\
\hline
\multirow{3}{*}{\makecell[l]{MSE\\ (N=3)}} & - & 0.9072 & 0.9322 & 0.9195 & 0.9317 & 0.8962 & 0.9136 & 0.9125 & 0.9280 & 0.9202 \\
                     & EI & 0.8864 & 0.9258 & 0.9057 & 0.9181 & 0.9025 & 0.9103 & 0.8975 & 0.9089 & 0.9032 \\
                     & LM & 0.8912 & 0.9195 & 0.9051 & 0.8987 & 0.9025 & 0.9006 & 0.9070 & 0.9301 & 0.9184 \\
                     
\hline
\multirow{3}{*}{\makecell[l]{ASE\\ (N=3)}} & - & 0.9091 & 0.9322 & 0.9205 & 0.9295 & 0.8941 & 0.9114 & 0.9146 & 0.9301 & \textbf{0.9223} \\
                     & EI & 0.8960 & 0.9131 & 0.9045 & 0.9124 & 0.9047 & 0.9085 & 0.9025 & 0.9025 & 0.9025 \\
                     & LM & 0.9021 & 0.9174 & 0.9097 & 0.8971 & 0.9047 & 0.9008 & 0.9160 & 0.9237 & 0.9198 \\
\hline
\end{tabular}
\caption{\label{table:ct-bert-results}
Result obtained for CT-BERT model with different fine-tuning strategies (All these experiments are executed for 5 learning epochs and S. abbreviates the Strategy)
}
\end{table*}

\subsection{Fine-tuning}
\label{subsec:fine_tuning}
To improve the models, we experimented different fine-tuning strategies: majority class self-ensemble, average self-ensemble, entity integration and language modelling, which are described below. 

\begin{enumerate}
\item{\textbf{Self-Ensemble (SE)}} - Self-ensemble is found as a technique which result better performance than the performance of a single model \cite{xu2020improving}. In this approach, same model architecture is trained or fine-tuned with different random seeds or train-validation splits. Then the output of each model is aggregated to generate the final results. As the aggregation methods, we analysed majority-class and average in this research. The number of models used with self-ensemble will be denoted by $N$.
\begin{itemize}
    \item \textit{Majority-class SE (MSE)} - As the majority class, we computed the mode of the classes predicted by each model. Given a data instance, following the softmax layer, a model predicts probabilities for each class and the class with highest probability is taken as the model predicted class. 

    \item \textit{Average SE (ASE)} - In average SE, final probability of class $c$ is calculated as the average of probabilities predicted by each model as in Equation \ref{equ:ase} where \textit{h} is the final hidden state of the [CLS] token. Then the class with highest probability is selected as the final class. 
    
    \begin{equation}
    \label{equ:ase}
    p_{ASE}(c|h) = \frac{\sum_{k=1}^{N} p_{k}(c|h) }{N}
    \end{equation}
\end{itemize}

\item{\textbf{Entity Integration (EI)}} - Since we are using pretrained models, there can be model unknown data in the task data set such as person names, locations and organisations. As entity integration, we replaced the unknown tokens with their named entities which are known to the model, so that the familiarity of data to model can be increased. To identify the named entities, we used the pretrained models available with spaCy \footnote{More details about spaCy are available on \url{https://spacy.io/}}.

\item{\textbf{Language Modelling (LM)}} - As language modelling, we retrained the transformer model on task data set before fine-tuning it for the downstream task; text classification. This training is took place according with the model's initial trained objective. Following this technique model understanding on the task data can be improved. 
\end{enumerate}

\section{Experiments and Results}
In this section, we report the experiments we conducted and their results. As informed by task organisers, we used precision, recall and F1 score calculated for \textit{Informative} class to measure the model performance. Results in sections \ref{res:tranformers} - \ref{res:fine-tuning} are computed on validation data set and results in section \ref{res:test-eval} are computed on test data set. 

\subsection{Impact by Transformer Model} \label{res:tranformers}
Initially we focused on the impact by different transformer models. Selected transformer models were fine-tuned for this task using single-model (no ensemble) and MSE with 3 models, and the obtained results are summarised in Table \ref{table:transformer-models}. According to the results, CT-BERT model outperformed the other models. Also, all the models except XLNet showed improved results with self-ensemble approach than single-model approach. Following these results and considering time and resource constraints, we limited the further experiments only to CT-BERT model. 

\subsection{Impact by Epoch Count} \label{res:epoch}
\begin{figure}[ht]
\centering
\includegraphics[scale=0.496]{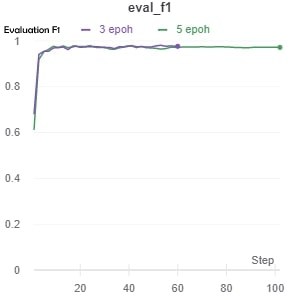}
\caption{Evaluation F1 score against the epoch count}
\label{fig:epoh_count}
\end{figure}

We experimented that increasing the epoch count from 3 to 5 increases the results. However, increasing it more than 5 did not further improved the results. Therefore, we used an epoch count of 5 in our experiments. To monitor the evaluation scores against the epoch count we used Wandb app \footnote{Wandb app is available on \url{https://app.wandb.ai/}}. As shown in the Figure \ref{fig:epoh_count} evaluation f1 score does not likely to change when trained with more than five epochs.

\subsection{Impact by Fine-tuning} \label{res:fine-tuning}
The fine-tuning strategies mentioned in Section \ref{subsec:fine_tuning} were experimented using CT-BERT model and obtained results are summarised in Table \ref{table:ct-bert-results}. According to the results, in majority of experiments, ASE is given a higher F1 than MSE. The other fine- strategies: EI and LM did not improve the results for this data set. As possible reasons for this reduction, having a good knowledge about COVID tweets by the model itself and insufficiency of data for language modelling can be mentioned.

Additionally, we analysed the impact by different learning rates. For initial experiments a random learning rate of $1e^{-5}$ was picked and for further analysis a less value ($1e^{-6}$) and a high value ($2e^{-5}$) were picked. The value $2e^{-5}$ was used for pretraining and experiments of CT-BERT model \cite{muller2020covid}. According to this analysis there is a tendency to have higher F1 with higher learning rates.

\subsection{Test Set Evaluation}\label{res:test-eval}
The test data results of our submissions, task baseline and top-ranked system are summarised in Table \ref{table:final-results}. Considering the evaluation results on validation data set, as InfoMiner 1 we selected the fine-tuned CT-BERT model with ASE and $2e^{-5}$ learning rate. As InfoMiner 2 same model and parameters with MSE was picked. Among them, the highest F1 we received is for MSE strategy.

\begin{table}[!htbp]
\centering
\begin{tabular}{l|l|l|l}
\hline
\textbf{Model} & \textbf{P} & \textbf{R} & \textbf{F1} \\
\hline
Top-ranked & 0.9135 & 0.9057 & 0.9096 \\
InfoMiner 1 & 0.9107 & 0.8856 & 0.8980 \\
InfoMiner 2 & 0.9102 & 0.8909 & 0.9004 \\
Task baseline & 0.7730 & 0.7288 & 0.7503 \\
\hline
\end{tabular}
\caption{\label{table:final-results}
Results of test data predictions
}
\end{table}

\section{Conclusion}
We have presented the system by InfoMiner team for WNUT-2020 Task 2. For this task, we have shown that the CT-BERT is the most successful transformer model from several transformer models we experimented. Furthermore, we presented several fine tuning strategies: self-ensemble, entity integration and language modelling that can improve the results. Overall, our approach is simple but can be considered as effective since it achieved 10$^{th}$ place in the leader-board.

As a future direction of this research, we hope to analyse the impact by different classification heads such as LSTM and Convolution Neural Network (CNN) in addition to softmax classifier on performance. Also, we hope to incorporate meta information-based features like number of retweets and likes with currently used textual features to involve social aspect for informative tweet identification.

\newpage
\bibliographystyle{acl_natbib}
\bibliography{emnlp2020}

\end{document}